\documentclass[11pt,a4paper]{article}
\usepackage{authblk}
\usepackage[hyperref]{emnlp-ijcnlp-2019}
\usepackage{times}
\usepackage{latexsym}

\usepackage{url}

\usepackage{graphicx}
\usepackage{amsmath}
\usepackage{booktabs}
\usepackage{algorithm}
\usepackage{algorithmic}
\usepackage{mathrsfs}
\usepackage{amssymb}
\usepackage{setspace}
\usepackage{comment}
\usepackage{xcolor}

\newtheorem{defn}{Definition}[section]

\aclfinalcopy 

\title{Meta Relational Learning for Few-Shot Link Prediction in Knowledge Graphs}

\author[1,*]{\textbf{Mingyang Chen}}
\author[1,*]{\textbf{Wen Zhang}}
\author[2]{\textbf{Wei Zhang}}
\author[2]{\textbf{Qiang Chen}}
\author[1,3,$\dagger$]{\textbf{Huajun Chen}}
\affil[1]{College of Computer Science and Technology, Zhejiang University}
\affil[2]{Alibaba Group}
\affil[3]{AZFT Joint Lab for Knowledge Engine}
\affil[ ]{\{mingyangchen, wenzhang2015, huajunsir\}@zju.edu.cn}
\affil[ ]{\{lantu.zw, lapu.cq\}@alibaba-inc.com}

\newcommand\blfootnote[1]{%
\begingroup 
\renewcommand\thefootnote{}\footnote{#1}%
\addtocounter{footnote}{-1}%
\endgroup 
}

\date{}

\begin{document}
\maketitle
\begin{abstract}

Link prediction is an important way to complete knowledge graphs (KGs), while embedding-based methods, effective for link prediction in KGs, perform poorly on relations that only have a few associative triples.
In this work, we propose a Meta Relational Learning (MetaR) framework to do the common but challenging few-shot link prediction in KGs, namely predicting new triples about a relation by only observing a few associative triples.
We solve few-shot link prediction by focusing on transferring relation-specific meta information to make model learn the most important knowledge and learn faster, corresponding to relation meta and gradient meta respectively in MetaR. 
Empirically, our model achieves state-of-the-art results on few-shot link prediction KG benchmarks.
\blfootnote{* \, Equal contribution.}
\blfootnote{$\dagger$ \, Corresponding author.}

\end{abstract}

\section{Introduction}
\label{sec:intro}

A knowledge graph is composed by a large amount of triples in the form of $(head\; entity,\, relation,\, tail\; entity)$ ($(h, r, t)$ in short), encoding knowledge and facts in the world.
Many KGs have been proposed \cite{wikidata, freebase, nell} and applied to various applications \cite{qa,recommender,longtail-re}.

Although with huge amount of entities, relations and triples, many KGs still suffer from incompleteness, thus knowledge graph completion is vital for the development of KGs. One of knowledge graph completion tasks is link prediction, predicting new triples based on existing ones. For link prediction, KG embedding methods \cite{TransE, RESCAL, ComplEx, DistMult} are promising ways. They learn latent representations, called embeddings, for entities and relations in continuous vector space and accomplish link prediction via calculation with embeddings. 

\begin{figure}[t]
\centering
\includegraphics[scale=0.41]{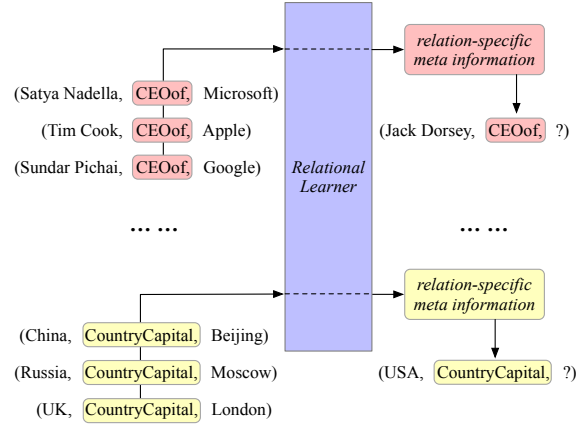}
\caption{An example of 3-shot link prediction in KGs. One task represents observing only three instances of one specific relation and conducting link prediction on this relation. Our model focuses on extracting relation-specific meta information by a kind of relational learner which is shared across tasks and transferring this meta information to do link prediction within one task.}
\label{fig:fsrl}
\end{figure}

The effectiveness of KG embedding methods is promised by sufficient training examples, thus results are much worse for elements with a few instances during training~\cite{crosse}. 
However, few-shot problem widely exists in KGs. For example, about 10\% of relations in Wikidata~\cite{wikidata} have no more than 10 triples. 
Relations with a few instances are called few-shot relations.
In this paper, we devote to discuss \emph{few-shot link prediction in knowledge graphs}, predicting tail entity $t$ given head entity $h$ and relation $r$ by only observing $K$ triples about $r$, usually $K$ is small.
Figure~\ref{fig:fsrl} depicts an example of 3-shot link prediction in KGs. 
 
To do few-shot link prediction, \citet{GMatching} made the first trial and proposed GMatching, learning a matching metric by considering both learned embeddings and one-hop graph structures, while we try to accomplish few-shot link prediction from another perspective based on the intuition that 
\emph{the most important information to be transferred from a few existing instances to incomplete triples should be the common and shared knowledge within one task.} 
We call such information \emph{relation-specific meta information} and propose a new framework Meta Relational Learning (MetaR) for few-shot link prediction. For example, in Figure~\ref{fig:fsrl}, relation-specific meta information related to the relation \emph{CEOof} or \emph{CountryCapital} will be extracted and transferred by MetaR from a few existing instances to incomplete triples.  

The relation-specific meta information is helpful in the following two perspectives: 1) transferring common relation information from observed triples to incomplete triples, 2) accelerating the learning process within one task by observing only a few instances. Thus we propose two kinds of relation-specific meta information: \emph{relation meta} and \emph{gradient meta} corresponding to afore mentioned two perspectives respectively. In our proposed framework MetaR, relation meta is the high-order representation of a relation connecting head and tail entities. 
Gradient meta is the loss gradient of relation meta which will be used to make a rapid update before transferring relation meta to incomplete triples during prediction.

Compared with GMatching~\cite{GMatching} which relies on a background knowledge graph, our MetaR is independent with them, thus is more robust as background knowledge graphs might not be available for few-shot link prediction in real scenarios. 

We evaluate MetaR with different settings on few-shot link prediction datasets. MetaR achieves state-of-the-art results, indicating the success of transferring relation-specific meta information in few-shot link prediction tasks.
In summary, main contributions of our work are three-folds:
\begin{itemize}
\item Firstly, we propose a novel meta relational learning framework (MetaR) to address few-shot link prediction in knowledge graphs.
\item Secondly, we highlight the critical role of relation-specific meta information for few-shot link prediction, and propose two kinds of relation-specific meta information, \emph{relation meta} and \emph{gradient meta}. Experiments show that both of them contribute significantly.
\item Thirdly, our MetaR achieves state-of-the-art results on few-shot link prediction tasks and we also analyze the facts that affect MetaR's performance. 
\end{itemize}

\section{Related Work}

One target of MetaR is to learn the representation of entities fitting the few-shot link prediction task and the learning framework is inspired by knowledge graph embedding methods. Furthermore, using loss gradient as one kind of meta information is inspired by MetaNet~\cite{metanet} and MAML~\cite{maml} which explore methods for few-shot learning by meta-learning. 
From these two points, we regard knowledge graph embedding and meta-learning as two main kinds of related work.

\subsection{Knowledge Graph Embedding}

Knowledge graph embedding models map relations and entities into continuous vector space. 
They use a score function to measure the truth value of each triple $(h, r, t)$. Same as knowledge graph embedding, our MetaR also need a score function, and the main difference is that representation for $r$ is the learned relation meta in MetaR rather than embedding of $r$ as in normal knowledge graph embedding methods.

One line of work is started by TransE \cite{TransE} with distance score function. TransH~\cite{TransH} and TransR~\cite{TransR} are two typical models using different methods to connect head, tail entities and their relations. DistMult~\cite{DistMult} and ComplEx~\cite{ComplEx} are derived from RESCAL~\cite{RESCAL}, trying to mine latent semantics in different ways. There are also some others like ConvE~\cite{conve} using convolutional structure to score triples and models using additional information such as entity types~\cite{entity-type} and relation paths~\cite{PTransE}.
\citet{kgembedding} comprehensively summarize the current popular knowledge graph embedding methods.

Traditional embedding models are heavily rely on rich training instances~\cite{iteratively, GMatching}, thus are limited to do few-shot link prediction. Our MetaR is designed to fill this vulnerability of existing embedding models.

\subsection{Meta-Learning}

Meta-learning seeks for the ability of learning quickly from only a few instances within the same concept and adapting continuously to more concepts, which are actually the rapid and incremental learning that humans are very good at.

Several meta-learning models have been proposed recently. Generally, there are three kinds of meta-learning methods so far: 
(1) \textit{Metric-based} meta-learning~\cite{siamese,matching,prototypical,GMatching}, which tries to learn a matching metric between query and support set generalized to all tasks, where the idea of matching is similar to some nearest neighbors algorithms. Siamese Neural Network~\cite{siamese} is a typical method using symmetric twin networks to compute the metric of two inputs. 
GMatching~\cite{GMatching}, the first trial on one-shot link prediction in knowledge graphs, learns a matching metric based on entity embeddings and local graph structures which also can be regarded as a metric-based method.
(2) \textit{Model-based} method~\cite{mann,metanet,snail}, which uses a specially designed part like memory to achieve the ability of learning rapidly by only a few training instances. MetaNet~\cite{metanet}, a kind of memory augmented neural network (MANN), acquires meta information from loss gradient and generalizes rapidly via its fast parameterization. 
(3) \textit{Optimization-based} approach~\cite{maml,gradient-based}, which gains the idea of learning faster by changing the optimization algorithm. Model-Agnostic Meta-Learning~\cite{maml} abbreviated as MAML is a model-agnostic algorithm. It firstly updates parameters of task-specific learner, and meta-optimization across tasks is performed over parameters by using above updated parameters, it's like ``a gradient through a gradient".

As far as we know, work proposed by \citet{GMatching} is the first research on few-shot learning for knowledge graphs. It's a metric-based model which consists of a neighbor encoder and a matching processor. Neighbor encoder enhances the embedding of entities by their one-hop neighbors, and matching processor performs a multi-step matching by a LSTM block. 

\begin{figure*}[t]
\centering
\includegraphics[scale=0.7]{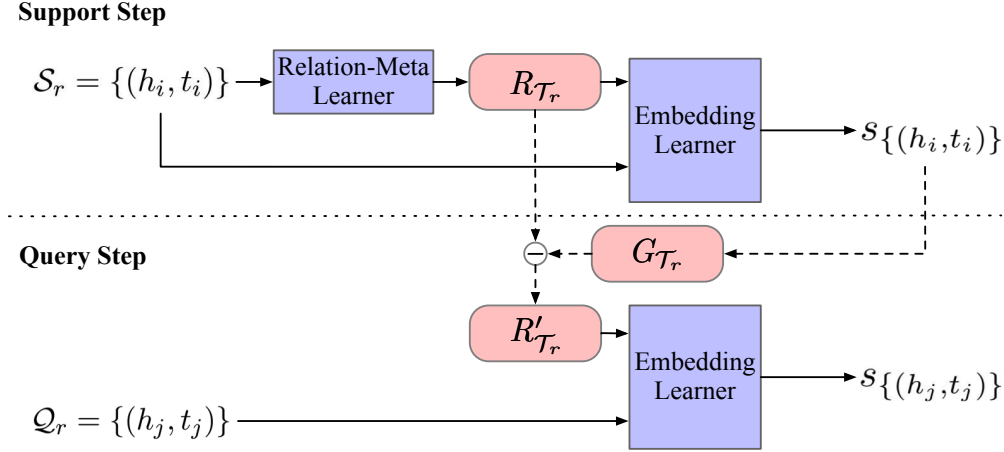}
\caption{Overview of MetaR. $\mathcal{T}_r = \{ \mathcal{S}_r, \mathcal{Q}_r\}$, $\mathit{R}_{\mathcal{T}_r}$ and $\mathit{R}_{\mathcal{T}_r}^{\prime}$ represent relation meta and updated relation meta, and $\mathit{G}_{\mathcal{T}_r}$ represents gradient meta.}
\vspace{-5mm}
\label{fig:metalp}
\end{figure*}

\section{Task Formulation}

\begin{table} 
\small
\centering
\setlength{\tabcolsep}{5mm}{
\begin{tabular}{l|l} 
\toprule
\multicolumn{2}{c}{Training} \\
\midrule
\multicolumn{2}{l}{Task \#1 ({\color{cyan} CountryCapital})} \\
\midrule
Support & (China, {\color{cyan} CountryCapital}, Beijing) \\
\midrule
Query & (France, {\color{cyan} CountryCapital}, Paris) \\
\midrule
\multicolumn{2}{l}{Task \#2 ({\color{red} CEOof})} \\
\midrule
Support & (Satya Nadella, {\color{red} CEOof}, Microsoft) \\
\midrule
Query & (Jack Dorsey, {\color{red} CEOof}, Twitter) \\
\midrule
\midrule
\multicolumn{2}{c}{Testing} \\
\midrule
\multicolumn{2}{l}{Task \#1 ({\color{blue} OfficialLanguage})} \\
\midrule
Support & (Japan, {\color{blue} OfficialLanguage}, Japanese)  \\
\midrule
Query & (Spain, {\color{blue} OfficialLanguage}, Spanish) \\

\bottomrule
\end{tabular}}
\caption{The training and testing examples of 1-shot link prediction in KGs.}
\label{tab:task-form}
\end{table}

In this section, we present the formal definition of a knowledge graph and few-shot link prediction task. A knowledge graph is defined as follows:
\begin{defn}
	(Knowledge Graph $\mathcal{G}$) A knowledge graph $\mathcal{G} = \{ \mathcal{E}, \mathcal{R},  \mathcal{TP}\}$. $\mathcal{E}$ is the entity set. $\mathcal{R}$ is the relation set. And $\mathcal{TP} = \{ (h, r, t)\in \mathcal{E} \times \mathcal{R} \times \mathcal{E}\} $ is the triple set. 
\end{defn}
And a few-shot link prediction task in knowledge graphs is defined as:
\begin{defn}
(Few-shot link prediction task $\mathcal{T}$) With a knowledge graph $\mathcal{G} = \{ \mathcal{E}, \mathcal{R},  \mathcal{TP}\}$, given a support set  $\mathcal{S}_r = \{(h_i, t_i)\in \mathcal{E} \times \mathcal{E} | (h_i, r, t_i) \in \mathcal{TP} \}$ about relation $r\in \mathcal{R}$, where $|\mathcal{S}_r | = K$, predicting the tail entity linked with relation $r$ to head entity $h_j$, formulated as $r:(h_j, ?)$, is called K-shot link prediction. 
\end{defn}

As defined above, a few-shot link prediction task is always defined for a specific relation. During prediction, there usually is more than one triple to be predicted, and  with \emph{support set} $\mathcal{S}_r$ , we call the set of all triples to be predicted as \emph{query set} $\mathcal{Q}_r = \{ r:(h_j, ?)\}$. 

The goal of a few-shot link prediction method is to gain the capability of predicting new triples about a relation $r$ with only observing a few triples about $r$. 
Thus its training process is based on a set of tasks $\mathcal{T}_{train}=\{\mathcal{T}_{i}\}_{i=1}^{M}$ where each task $\mathcal{T}_{i} = \{\mathcal{S}_i, \mathcal{Q}_i\}$ corresponds to an individual few-shot link prediction task with its own support and query set.
Its testing process is conducted on a set of new tasks $\mathcal{T}_{test} = \{\mathcal{T}_{j}\}_{j=1}^{N}$ which is similar to $\mathcal{T}_{train}$, other than that $\mathcal{T}_{j} \in \mathcal{T}_{test}$ should be about relations that have never been seen in $\mathcal{T}_{train}$.

Table~\ref{tab:task-form} gives a concrete example of the data during learning and testing for few-shot link prediction.

\section{Method}

To make one model gain the few-shot link prediction capability, the most important thing is transferring information from support set to query set and there are two questions for us to think about: 
(1) what is the most transferable and common information between support set and query set 
and (2) how to learn faster by only observing a few instances within one task. 
For question (1), within one task, all triples in support set and query set are about the same relation, thus it is naturally to suppose that relation is the key common part between support and query set.  For question (2), the learning process is usually conducted by minimizing a loss function via gradient descending, thus gradients reveal how the model's parameters should be changed. Intuitively, we believe that gradients are valuable source to accelerate learning process. 

Based on these thoughts, we propose two kinds of meta information which are shared between support set and query set to deal with above problems:

\begin{itemize}
\item \textbf{Relation Meta} represents the relation connecting head and tail entities in both support and query set and we extract relation meta for each task, represented as a vector, from support set and transfer it to query set.

\item \textbf{Gradient Meta} is the loss gradient of relation meta in support set. As gradient meta shows how relation meta should be changed in order to reach a loss minima, thus to accelerate the learning process, relation meta is updated through gradient meta before being transferred to query set. This update can be viewed as the rapid learning of relation meta.
\end{itemize}

In order to extract relation meta and gradient mate and incorporate them with knowledge graph embedding to solve few-shot link prediction, our proposal, \textbf{MetaR}, mainly contains two modules:

\begin{itemize}
\item \textbf{Relation-Meta Learner} generates relation meta from heads' and tails' embeddings in the support set.
\item \textbf{Embedding Learner} calculates the truth values of triples in support set and query set via entity embeddings and relation meta. Based on the loss function in embedding learner, gradient meta is calculated and a rapid update for relation meta will be implemented before transferring relation meta to query set.
\end{itemize}

The overview and algorithm of MetaR are shown in Figure~\ref{fig:metalp} and Algorithm~\ref{alg:metalp}. Next, we introduce each module of MetaR via one few-shot link prediction task $\mathcal{T}_r = \{ \mathcal{S}_r, \mathcal{Q}_r\}$. 

\begin{algorithm}[tb]
\setstretch{1}
\caption{Learning of MetaR}
\label{alg:metalp}
\begin{algorithmic}[1]
\REQUIRE Training tasks $\mathcal{T}_{train}$
\REQUIRE Embedding layer $emb$; Parameter of relation-meta learner $\phi$
\WHILE{not done}
\STATE Sample a task $\mathcal{T}_r={\{\mathcal{S}_r, \mathcal{Q}_r\}}$ from $\mathcal{T}_{train}$
\STATE Get $\mathit{R}$ from $\mathcal{S}_{r}$ (Equ.~\ref{eq:rel-meta}, Equ.~\ref{eq:rel-avg})
\STATE Compute loss in $\mathcal{S}_{r}$ (Equ.~\ref{eq:loss-sup})
\STATE Get $\mathit{G}$ by gradient of $\mathit{R}$ (Equ.~\ref{eq:grad-meta})
\STATE Update $\mathit{R}$ by $\mathit{G}$ (Equ.~\ref{eq:update-rel-meta})
\STATE Compute loss in $\mathcal{Q}_{r}$ (Equ.~\ref{eq:loss-que})
\STATE Update $\phi$ and $emb$ by loss in $\mathcal{Q}_{r}$
\ENDWHILE

\end{algorithmic}
\end{algorithm}

\subsection{Relation-Meta Learner}

To extract the relation meta from support set, we design a relation-meta learner to learn a mapping from head and tail entities in support set to relation meta. The structure of this relation-meta learner can be implemented as a simple neural network. 

In task $\mathcal{T}_r$, the input of relation-meta learner is head and tail entity pairs in support set $\{(h_i, t_i) \in \mathcal{S}_r\}$. We firstly extract entity-pair specific relation meta via a $L$-layers fully connected neural network,

\begin{equation}
\begin{aligned}
  \mathbf{x}^0 &= \mathbf{h}_i \oplus \mathbf{t}_i \\
  \mathbf{x}^l &= \sigma({\mathbf{W}^{l}\mathbf{x}^{l-1} + b^l}) \\
  \mathit{R}_{(h_i, t_i)} &= {\mathbf{W}^{L}\mathbf{x}^{L-1} + b^L}
\label{eq:rel-meta}
\end{aligned}
\end{equation}

where $\mathbf{h}_i \in \mathbb{R}^{d}$ and $\mathbf{t}_i  \in \mathbb{R}^{d}$ are embeddings of head entity $h_i$ and tail entity $t_i$ with dimension $d$ respectively. $L$ is the number of layers in neural network, and $l \in \{1, \dots, L-1 \}$. $\mathbf{W}^l$ and $\mathbf{b}^l$ are weights and bias in layer $l$. We use LeakyReLU for activation $\sigma$. $\mathbf{x} \oplus \mathbf{y}$ represents the concatenation of vector $\mathbf{x}$ and $\mathbf{y}$. Finally, $\mathit{R}_{(h_i, t_i)}$ represent the relation meta from specific entity pare $h_i$ and $t_i$.
 
With multiple entity-pair specific relation meta, we generate the final relation meta in current task via averaging all entity-pair specific relation meta in current task,

\begin{equation}
\mathit{R}_{\mathcal{T}_r} = \frac{\sum_{i=1}^{K}\mathit{R}_{(h_i, t_i)}}{K}
\label{eq:rel-avg}
\end{equation}

\subsection{Embedding Learner}


As we want to get gradient meta to make a rapid update on relation meta, we need a score function to evaluate the truth value of entity pairs under specific relations and also the loss function for current task. We apply the key idea of knowledge graph embedding methods in our embedding learner, as they are proved to be effective on evaluating truth value of triples in knowledge graphs.     

In task $\mathcal{T}_r$, we firstly calculate the score for each entity pairs $(h_i, t_i)$ in support set $\mathcal{S}_r$ as follows:
 \begin{equation}
  	s_{(h_i, t_i)} = \| \mathbf{h}_i + {\mathit{R}_{\mathcal{T}_r}} - \mathbf{t}_i \| 
\label{eq:score-sup}
\end{equation}
where $\| \mathbf{x}\|$ represents the L2 norm of vector $\mathbf{x}$. We design the score function inspired by TransE \cite{TransE} which assumes the head entity embedding $\mathbf{h}$, relation embedding $\mathbf{r}$ and tail entity embedding $\mathbf{t}$ for a true triple $(h, r, t)$ satisfying $\mathbf{h} + \mathbf{r} = \mathbf{t}$. Thus the score function is defined according to the distance between $\mathbf{h} + \mathbf{r} $ and $\mathbf{t}$. Transferring to our few-show link prediction task, we replace the relation embedding $\mathbf{r}$ with relation meta $\mathit{R}_{\mathcal{T}_r}$ as there is no direct general relation embeddings in our task and $\mathit{R}_{\mathcal{T}_r}$ can be regarded as the relation embedding for current task $\mathcal{T}_r$. 

With score function for each triple, we set the following loss,
\begin{equation}
	L(\mathcal{S}_r) = \sum_{(h_i, t_i)\in \mathcal{S}_r} [\gamma+s_{(h_i, t_i)}-s_{(h_i, t_i^{\prime})}]_{+}
\label{eq:loss-sup}
\end{equation}
where $[x]_{+}$ represents the positive part of $x$ and $\gamma$ represents margin which is a hyperparameter. $s_{(h_i, t_i^{\prime})}$ is the score for negative sample $(h_i, t_i^{\prime})$ corresponding to current positive entity pair $(h_i, t_i) \in \mathcal{S}_r$, where $(h_i, r, t_i^{\prime}) \notin \mathcal{G}$.


$L(\mathcal{S}_r)$ should be small for task $\mathcal{T}_r$ which represents the model can properly encode truth values of triples. Thus gradients of parameters indicate how should the parameters be updated. Thus we regard the gradient of $\mathit{R}_{\mathcal{T}_r}$ based on $L(\mathcal{S}_r)$ as gradient meta $\mathit{G}_{\mathcal{T}_r}$:
\begin{equation}
\vspace{-1mm}
  \mathit{G}_{\mathcal{T}_r} = \nabla_{\mathit{R}_{\mathcal{T}_r}} L(\mathcal{S}_r)
\label{eq:grad-meta}
\end{equation}
Following the gradient update rule, we make a rapid update on relation meta as follows:
\begin{equation}
  \mathit{R}^\prime_{\mathcal{T}_r} = \mathit{R}_{\mathcal{T}_r} - \beta \mathit{G}_{\mathcal{T}_r}
\label{eq:update-rel-meta}
\end{equation}
where $\beta$ indicates the step size of gradient meta when operating on relation meta.

When scoring the query set by embedding learner, we use updated relation meta. 
After getting the updated relation meta $\mathit{R}^\prime$, we transfer it to samples in query set $\mathcal{Q}_r = \{(h_j, t_j) \}$ and calculate their scores and loss of query set, following the same way in support set:
\begin{equation}
  s_{(h_j, t_j)} = \| \mathbf{h}_j + \mathit{R}_{\mathcal{T}_r}^\prime - \mathbf{t}_j \| 
\label{eq:score-que}
\end{equation}
\begin{equation}
	L(\mathcal{Q}_r) = \sum_{(h_j, t_j)\in \mathcal{Q}_r}[\gamma+s_{(h_j, t_j)}-s_{(h_j, t_j^{\prime})}]_{+}
\label{eq:loss-que}
\end{equation}
where $L(\mathcal{Q}_r)$ is our training objective to be minimized. We use this loss to update the whole model.

\subsection{Training Objective}
During training, our objective is to minimize the following loss $L$ which is the sum of query loss  for all tasks in one minibatch: 
\begin{equation}
	L = \sum_{(\mathcal{S}_r, \mathcal{Q}_r)\in \mathcal{T}_{train}} L(\mathcal{Q}_r)
\end{equation}

\section{Experiments}

With MetaR, we want to figure out following things:
1) can MetaR accomplish few-shot link prediction task and even perform better than previous model? 
2) how much relation-specific meta information contributes to few-shot link prediction?
3) is there any requirement for MetaR to work on few-shot link prediction? 
To do these, we conduct the experiments on two few-shot link prediction datasets and deeply analyze the experiment results \footnote{The source code of experiments is available at \url{https://github.com/AnselCmy/MetaR}}. 

\begin{table} 
\centering
\begin{tabular}{ccccc}  
\toprule
Dataset & Fit & \# Train & \# Dev & \# Test \\
\midrule
NELL-One & Y & 321 & 5 & 11 \\
Wiki-One & Y & 589 & 16 & 34 \\
\midrule
NELL-One & N & 51 & 5 & 11 \\
Wiki-One & N & 133 & 16 & 34  \\
\bottomrule
\end{tabular}
\caption{Statistic of datasets. Fit denotes fitting background into training tasks (Y) or not (N), \# Train, \# Dev and \# Test denote the number of relations in training, validation and test set.}
\label{tab:statistic}
\end{table}

\subsection{Datasets and Evaluation Metrics}
\label{sec:dataset}

We use two datasets, NELL-One and Wiki-One which are constructed by~\citet{GMatching}. NELL-One and Wiki-One are derived from NELL~\cite{nell} and Wikidata~\cite{wikidata} respectively. Furthermore, because these two benchmarks are firstly tested on GMatching which consider both learned embeddings and one-hop graph structures, a background graph is constructed with relations out of training/validation/test sets for obtaining the pre-train entity embeddings and providing the local graph for GMatching.

Unlike GMatching using background graph to enhance the representations of entities, our MetaR can be trained without background graph. For NELL-One and Wiki-One which have background graph originally, we can make use of such background graph by fitting it into training tasks or using it to train embeddings to initialize entity representations. Overall, we have three kinds of dataset settings, shown in Table~\ref{tab:dataform}. For setting of \emph{BG:In-Train}, in order to make background graph included in training tasks, we sample tasks from triples in background graph and original training set, rather than sampling from only original training set.

\begin{table} 
\centering
\begin{tabular}{cp{110pt}} 
\toprule
Background Conf. & Description \\ 
\midrule
BG:Pre-Train & Use background to train entity embedding in advance.\\  
\midrule 
BG:In-Train & Fit background graph into training tasks. \\ 
\midrule
BG:Discard & Discard the background graph. \\ 
\bottomrule
\end{tabular}
\caption{Three forms of datasets in our experiments.}
\label{tab:dataform}
\end{table}

Note that these three settings don't violate the task formulation of few-shot link prediction in KGs. The statistics of NELL-One and Wiki-One are shown in Table~\ref{tab:statistic}.

We use two traditional metrics to evaluate different methods on these datasets, MRR and Hits@N. MRR is the mean reciprocal rank and Hits@N is the proportion of correct entities ranked in the top N in link prediction.

\begin{table*}[t]
\centering
\begin{tabular}{lcc|cc|cc|cc}  
\toprule

\multicolumn{1}{c}{} & \multicolumn{2}{c}{MRR} & \multicolumn{2}{c}{Hits@10} & \multicolumn{2}{c}{Hits@5} & \multicolumn{2}{c}{Hits@1}\\ 
\cmidrule{2-9} 
\textbf{NELL-One} & 1-shot & 5-shot & 1-shot & 5-shot & 1-shot & 5-shot & 1-shot & 5-shot \\

\midrule
GMatching\_RESCAL & \underline{.188} & -- 
          			& .305 & -- 
          			& .243 & -- 
          			& \underline{.133} & -- \\ 
GMatching\_TransE & .171 & -- 
          			& .255 & -- 
          			& .210 & -- 
          			& .122 & -- \\ 
GMatching\_DistMult & .171 & -- 
          			& .301 & -- 
          			& .221 & -- 
          			& .114 & -- \\ 
GMatching\_ComplEx & .185 & \underline{.201} 
          			& \underline{.313} & \underline{.311} 
          			& \underline{.260} & \underline{.264} 
          			& .119 & \underline{.143} \\ 
GMatching\_Random & .151 & -- 
          			& .252 & -- 
          			& .186 & -- 
          			& .103 & -- \\ 
\midrule
MetaR (BG:Pre-Train) & .164 & .209
							& .331 & .355
							& .238 & .280
							& .093 & .141 \\
MetaR (BG:In-Train) & \textbf{.250} & \textbf{.261} 
							& \textbf{.401} & \textbf{.437}
							& \textbf{.336} & \textbf{.350}
							& \textbf{.170} & \textbf{.168} \\
\midrule
\midrule
\textbf{Wiki-One} & 1-shot & 5-shot & 1-shot & 5-shot & 1-shot & 5-shot & 1-shot & 5-shot \\ 
\midrule
GMatching\_RESCAL & .139 & -- 
          			& .305 & -- 
          			& .228 & -- 
          			& .061 & -- \\ 
GMatching\_TransE & .219 & -- 
          			& .328 & -- 
          			& .269 & -- 
          			& .163 & -- \\ 
GMatching\_DistMult & \underline{.222} & -- 
          			& \underline{.340} & -- 
          			& .271 & -- 
          			& \underline{.164} & -- \\ 
GMatching\_ComplEx & .200 & -- 
          			& .336 & -- 
          			& \underline{.272} & -- 
          			& .120 & -- \\ 
GMatching\_Random & .198 & -- 
          			& .299 & -- 
          			& .260 & -- 
          			& .133 & -- \\ 
\midrule
MetaR (BG:Pre-Train) & \textbf{.314} & \textbf{.323}
							& \textbf{.404} & \textbf{.418}
							& \textbf{.375} & \textbf{.385}
							& \textbf{.266} & \textbf{.270}\\
MetaR (BG:In-Train) & .193 & .221
							& .280 & .302
							& .233 & .264
							& .152 & .178 \\
\bottomrule
\end{tabular}
\caption{Results of few-shot link prediction in NELL-One and Wiki-One. \textbf{Bold} numbers are the best results of all and \underline{underline} numbers are the best results of GMatching. The contents of (bracket) after MetaR illustrate the form of datasets we use for MetaR.}
\label{tab:result-fsrl}
\end{table*}

\subsection{Implementation}

During training, mini-batch gradient descent is applied with batch size set as 64 and 128 for NELL-One and Wiki-One respectively. We use Adam~\cite{adam} with the initial learning rate as 0.001 to update parameters. We set $\gamma = 1$ and $\beta = 1$. The number of positive and negative triples in query set is 3 and 10 in NELL-One and Wiki-One. Trained model will be applied on validation tasks each 1000 epochs, and the current model parameters and corresponding performance will be recorded, after stopping, the model that has the best performance on Hits@10 will be treated as final model. For number of training epoch, we use early stopping with 30 patient epochs, which means that we stop the training when the performance on Hits@10 drops 30 times continuously. Following GMatching, the embedding dimension of NELL-One is 100 and Wiki-One is 50. 
The sizes of two hidden layers in relation-meta learner are 500, 200 and 250, 100 for NELL-One and Wiki-One.


\subsection{Results}

The results of two few-shot link prediction tasks, including 1-shot and 5-shot, on NELL-One and Wiki-One are shown in Table~\ref{tab:result-fsrl}. 
The baseline in our experiment is GMatching~\cite{GMatching}, which made the first trial on few-shot link prediction task and is the only method that we can find as baseline. 
In this table, results of GMatching with different KG embedding initialization are copied from the original paper.
Our MetaR is tested on different settings of datasets introduced in Table~\ref{tab:dataform}.

In Table~\ref{tab:result-fsrl}, our model performs better with all evaluation metrics on both datasets. 
Specifically, for 1-shot link prediction, MetaR increases by 33\%, 28.1\%, 29.2\% and 27.8\% on MRR, Hits@10, Hits@5 and Hits@1 on NELL-One, and 41.4\%, 18.8\%, 37.9\% and 62.2\% on Wiki-One, with average improvement of 29.53\% and 40.08\% respectively.
For 5-shot, MetaR increases by 29.9\%, 40.5\%, 32.6\% and 17.5\% on MRR, Hits@10, Hits@5 and Hits@1 on NELL-One with average improvement of 30.13\%.

Thus for the first question we want to explore, the results of MetaR are no worse than GMatching, indicating that MetaR has the capability of accomplishing few-shot link prediction.
In parallel, the impressive improvement compared with GMatching demonstrates that the key idea of MetaR, transferring relation-specific meta information from support set to query set, works well on few-shot link prediction task.  

Furthermore, compared with GMatching, our MetaR is independent with background knowledge graphs. We test MetaR on 1-shot link prediction in partial NELL-One and Wiki-One which discard the background graph, and get the results of 0.279 and 0.348 on Hits@10 respectively. Such results are still comparable with GMatching in fully datasets with background.

\begin{table} 
\centering
\begin{tabular}{lccc}  
\toprule
Ablation Conf. & BG:Pre-Train & BG:In-Train \\
\midrule
\textit{standard} & .331 & .401 \\
\midrule
\textit{-g} & .234 & .341 \\
\midrule
\textit{-g -r} & .052 & .052\\
\bottomrule
\end{tabular}
\caption{Results of ablation study on Hits@10 of 1-shot link prediction in NELL-One.}
\label{tab:result-ablation}
\end{table}

\subsection{Ablation Study}
We have proved that relation-specific meta information, the key point of MetaR, successfully contributes to few-shot link prediction in previous section. 
As there are two kinds of relation-specific meta information in this paper, relation meta and gradient meta, we want to figure out how these two kinds of meta information contribute to the performance. Thus, we conduct an ablation study with three settings. 
The first one is our complete MetaR method denoted as \textit{standard}. 
The second one is removing the gradient meta by transferring un-updated relation meta directly from support set to query set without updating it via gradient meta, denoted as \textit{-g}.
The third one is removing the relation meta further which makes the model rebase to a simple TransE embedding model, denoted as \textit{-g -r}. The result under the third setting is copied from \citet{GMatching}. It uses the triples from background graph, training tasks and one-shot training triples from validation/test set, so it's neither \textit{BG:Pre-Train} nor \textit{BG:In-Train}. 
We conduct the ablation study on NELL-one with metric Hit@10 and results are shown in Table~\ref{tab:result-ablation}. 


Table~\ref{tab:result-ablation} shows that removing gradient meta decreases 29.3\% and 15\% on two dataset settings,  
and further removing relation meta continuous decreases the performance with 55\% and 72\% compared to the \emph{standard} results. 
Thus both relation meta and gradient meta contribute significantly and relation meta contributes more than gradient meta. 
Without gradient meta and relation meta, there is no relation-specific meta information transferred in the model and it almost doesn't work. This also illustrates that relation-specific meta information is important and effective for few-shot link prediction task. 

\subsection{Facts That Affect MetaR's Performance}
\label{sec:analysis}

We have proved that both relation meta and gradient meta surely contribute to few-shot link prediction. But is there any requirement for MetaR to ensure the performance on few-shot link prediction? We analyze this from two points based on the results, one is the sparsity of entities and the other is the number of tasks in training set.

\textbf{The sparsity of entities} We notice that the best result of NELL-One and Wiki-One appears in different dataset settings. With NELL-One, MetaR performs better on \textit{BG:In-Train} dataset setting, while with Wiki-One, it performs better on \textit{BG:Pre-Train}. Performance difference between two dataset settings is more significant on Wiki-One. 

Most datasets for few-shot task are sparse and the same with NELL-One and Wiki-One, but the entity sparsity in these two datasets are still significantly different, which is especially reflected in the proportion of entities that only appear in one triple in training set, $82.8$\% and $37.1$\% in Wiki-One and NELL-One respectively. Entities only have one triple during training will make MetaR unable to learn good representations for them, because entity embeddings heavily rely on triples related to them  in MetaR. Only based on one triple, the learned entity embeddings will include a lot of bias. Knowledge graph embedding method can learn better embeddings than MetaR for those one-shot entities, because entity embeddings can be corrected by embeddings of relations that connect to it, while they can't in MetaR. This is why the best performance occurs in \textit{BG:Pre-train} setting on Wiki-One, pre-train entity embeddings help MetaR overcome the low-quality on one-shot entities.

\textbf{The number of tasks}
From the comparison of MetaR's performance between with and without background dataset setting on NELL-One, we find that the number of tasks will affect MetaR's performance significantly. 
With \textit{BG:In-Train}, there are 321 tasks during training and MetaR achieves 0.401 on Hits@10, while without background knowledge, there are 51, with 270 less, and MetaR achieves 0.279. 
This makes it reasonable that why MetaR achieves best performance on \textit{BG:In-Train} with NELL-One. Even NELL-One has $37.1$\% one-shot entities, adding background knowledge into dataset increases the number of training tasks significantly, which complements the sparsity problem and contributes more to the task.  

Thus we conclude that both the sparsity of entities and number of tasks will affect performance of MetaR. Generally, with more training tasks, MetaR performs better and for extremely sparse dataset, pre-train entity embeddings are preferred. 

\section{Conclusion}

We propose a meta relational learning framework to do few-shot link prediction in KGs, and we design our model to transfer relation-specific meta information from support set to query set. Specifically, using relation meta to transfer common and important information, and using gradient meta to accelerate learning. Compared to GMatching which is the only method in this task, our method MetaR gets better performance and it is also independent with background knowledge graphs. Based on experimental results, we analyze that the performance of MetaR will be affected by the number of training tasks and sparsity of entities. We may consider obtaining more valuable information about sparse entities in few-shot link prediction in KGs in the future.

\section*{Acknowledgments}

We want to express gratitude to the anonymous reviewers for their hard work and kind comments, which will further improve our work in the future. This work is funded by NSFC 91846204/61473260, national key research program YS2018YFB140004, and Alibaba CangJingGe(Knowledge Engine) Research Plan.

\bibliography{emnlp-ijcnlp-2019}
\bibliographystyle{acl_natbib}

\end{document}